\documentclass[letterpaper,10pt,conference]{ieeeconf}

\IEEEoverridecommandlockouts
\usepackage{amsmath,amssymb,amsfonts}
\usepackage{algorithmic}
\usepackage{graphicx}
\usepackage{textcomp}
\usepackage{xcolor}
\usepackage{url}
\usepackage{hyperref}
\usepackage{cite}

\newcommand{\cameraimage}{I^\mathrm{RGB-D}}

\usepackage{booktabs}
\usepackage{makecell}

\usepackage[frozencache]{minted}

\begin{document}
\bstctlcite{BSTcontrol}

\title{VLMs Can Describe, But Not Measure: Object-Centric Scene Understanding for Robotic Manipulation}

\author{
    Enrico Saccon$^\dagger$, Tommaso Faraci$^\dagger$, I\~{n}igo De La Ossa Zarzuelo, Luigi Palopoli, Marco Roveri, Matteo Saveriano
    \thanks{
        Enrico Saccon, Tommaso Faraci, Luigi Palopoli and Marco Roveri are with the Department of Information Engineering and Computer Science, the University of Trento, Trento, Italy. \{name.surname\}@unitn.it.}
    \thanks{
        I\~{n}igo De La Ossa Zarzuelo is with the Polytechnic School, Mondrag\'{o}n University, Mondrag\'{o}n, Spain. inigodelaossa@gmail.com }
    \thanks{
        Matteo Saveriano is with the Department of Industrial Engineering, University of Trento, Trento, Italy. matteo.saveriano@unitn.it.}
    \thanks{Enrico Saccon and Tommaso Faraci equally contributed to this work.}
    \thanks{Co-funded by the European Union under the INVERSE project (Grant Agreement No. 101136067).}
    \thanks{The dataset is publicly available~\cite{dataset}. The code of the pipeline is shared at \url{https://www.github.com/idra-lab/plantorv}.}
    \thanks{This work has been submitted to the IEEE for possible publication. Copyright may be transferred without notice, after which this version may no longer be accessible.}
}

\maketitle

\begin{abstract}
Robotic operation in previously unseen environments requires both semantic understanding and reliable metric information. While vision--language models (VLMs) provide strong semantic capabilities, their geometric estimates remain less reliable. In this paper, we propose a VLM-driven, modular perception framework for scene understanding using off-the-shelf approaches. 
Starting from a single RGB-D observation, the scene is segmented into object-level regions, annotated by a VLM, and grounded with depth information to construct a task-independent object-centric representation. Experiments on 151 tabletop scenes show that the proposed decomposition preserves strong semantic performance while substantially improving localization and depth estimation over direct VLM inference. The resulting representation is also integrated with a task-planning framework for robotic execution.
\end{abstract}


\section{Introduction}\label{sec:introduction}

Robotic manipulation in previously unseen environments is a fundamental open-issue, limiting the widespread of robotic solutions in unstructured environments. 
A robotic manipulator must recover semantic properties, metric geometry, and spatial relationships in a form that can subsequently support task reasoning and planning. Recent vision-language models (VLMs) \cite{singh2026openaigpt5card,lian2025describe} provide strong open-vocabulary semantic understanding without task-specific training, but are less suited to recovering the precise metric information required for physical interaction. Conversely, geometric perception methods \cite{depthanything3,wang2025vggt} can recover the three-dimensional structure of a scene, but generally provide much weaker semantic understanding. As a consequence, most systems remain designed a for specific task and a single environment, and fail to address greater autonomy in unseen settings, which would require robots to perceive and reason about the environment.

\begin{figure*}[t]
    \centering
    \includegraphics[width=\textwidth]{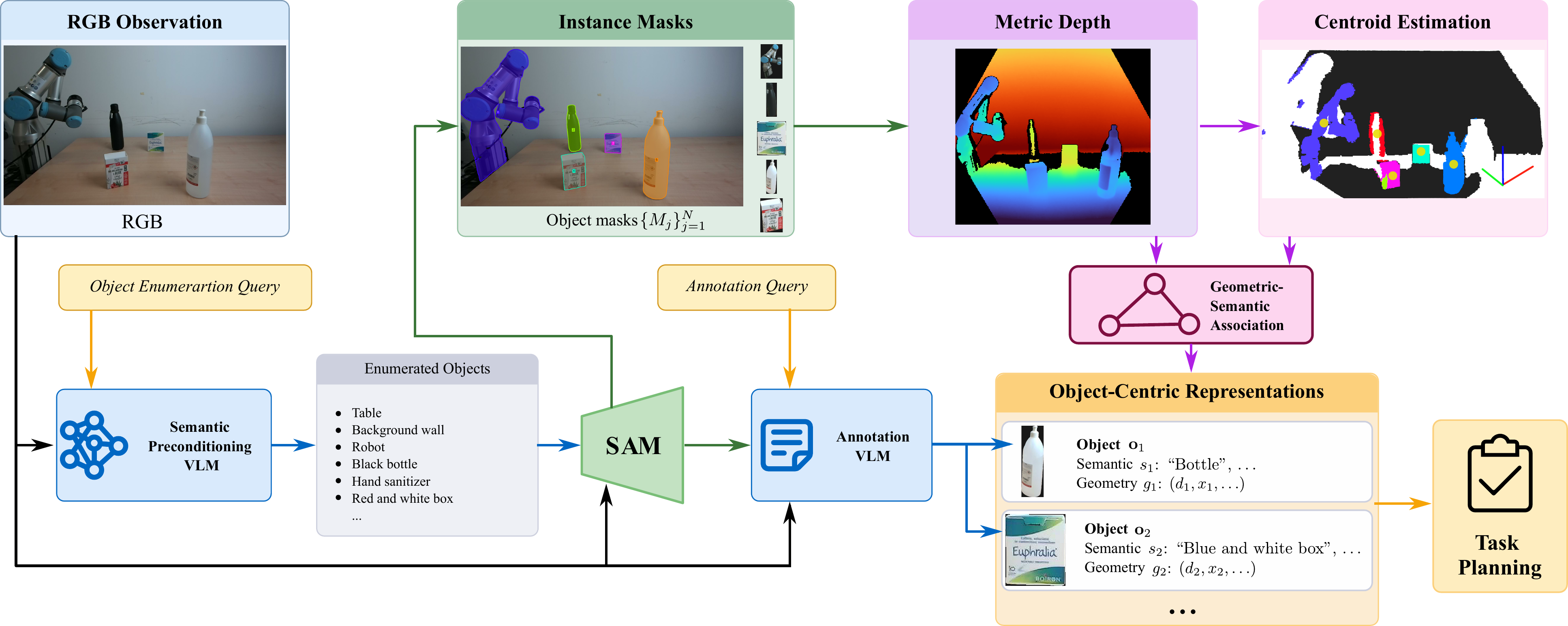}
    \caption{Overview of the proposed scene-understanding pipeline. From a single RGB-D observation, object candidates are segmented and semantically annotated using VLMs, associated with metric depth to estimate object centroids, and fused into an object-centric geometric–semantic representation for downstream task planning.}
    \label{fig:diagram}
\end{figure*}

In this work, we investigate whether these complementary capabilities can be combined to obtain a {task-independent description of an unknown scene from a single RGB-D observation, without scene-specific training or fine-tuning}. We explicitly separate semantic interpretation (VLM-driven) from geometric estimation (fusing segmentation and depth information), using \textit{off-the-shelf models}. In our framework (Fig.~\ref{fig:diagram}), foundation segmentation models such as SAM~\cite{carion2025sam3segmentconcepts} 
first decompose the RGB observation into candidate scene entities. The complete image and the resulting object masks are then annotated by a general-purpose, multimodal foundation model~\cite{singh2026openaigpt5card,lian2025describe}
. Finally, the segmented regions are associated with the corresponding RGB-D measurements to recover metric object information. We empirically demonstrate the effectiveness of this approach compared to end-to-end VLM baselines (Sec.~\ref{ssec:vlmBaseline}--\ref{ssec:ablation}).

\begin{figure}[t]
    \centering
    \includegraphics[width=\linewidth]{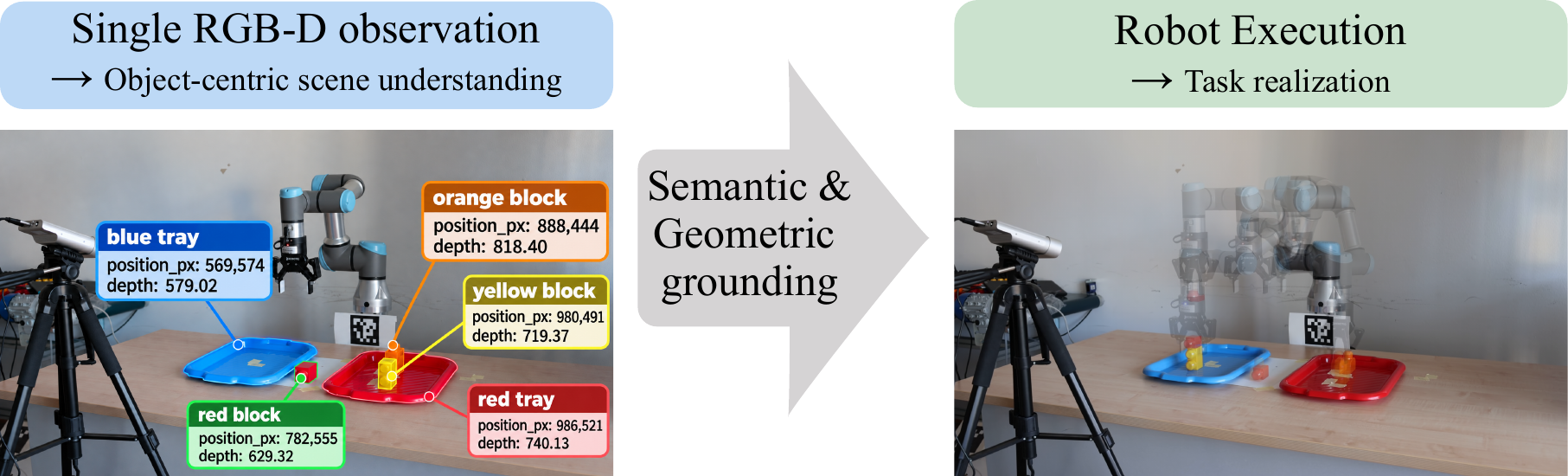}
    \caption{A single RGB-D observation is converted into a semantically and geometrically grounded object-centric representation (left) and subsequently used for robotic task execution (right); ghosted poses depict the manipulation sequence.}
    \label{fig:summary}
\end{figure}

The resulting object-centric representation combines semantic and geometric information in a form that can be queried independently of the task. It can therefore serve as an interface to symbolic and language-based planning frameworks such as~\cite{saccon2025automated,saccon2026combininglargelanguagemodels,gestrin2025nl2planrobustllmdrivenplanning}, linking open-vocabulary perception with downstream robotic planning. We demonstrate the effectiveness of this integration in a real robot sorting task (Sec.~\ref{ssec:realWorldDemonstration}), shown in Fig.~\ref{fig:summary}.

\section{Related Work}\label{sec:background}

\subsection{Learning and Foundation Models for Robotic Manipulation}


Learning-based robotic manipulation has increasingly exploited heterogeneous multimodal supervision. M$\pi$X~\cite{shah2023mutex} learns manipulation policies from demonstrations expressed through speech, images, text, and video through modality-specific encoders and a unified training framework. SAILOR~\cite{nasiriany2022sailor} similarly leverages heterogeneous offline data for imitation learning, while BridgeData~\cite{walke2023bridgedata} provides large-scale manipulation trajectories across diverse environments. These approaches improve behavioral generalization but primarily learn observation-to-action mappings from previously collected robot data rather than constructing an explicit, reusable representation of an unseen scene.


Language-conditioned methods, instead, combine semantic reasoning with geometric information. CLIPort~\cite{shridhar2021cliportpathwaysroboticmanipulation} integrates language-conditioned semantic features with RGB-D observations to learn language-conditioned pick-and-place policies. VoxPoser~\cite{huang2023voxposercomposable3dvalue} and ReKep~\cite{huang2024rekepspatiotemporalreasoningrelational} ground high-level reasoning into 3D spatial and temporal manipulation. These methods show the importance of combining high-level semantic reasoning with geometric information. However, their representations are task-conditioned, focusing on the entities and relations required for a specific instruction. In contrast, our work focuses on constructing a task-independent representation.

\subsection{Geometric and Open-Vocabulary Scene Understanding}


A complementary line of research focuses on geometry reconstruction. Modern monocular models such as Depth Anything~\cite{depthanything3} 
 can estimate dense depth maps from a single RGB image, while models such as VGGT~\cite{wang2025vggt} exploit learned geometric priors for 3D reconstruction and camera estimation. Nevertheless, single-view reconstruction remains intrinsically ambiguous, and predicted depth is not necessarily metric. Multi-view observations can alleviate these limitations, but are incompatible with settings with single observations. RGB-D sensing, instead, provides direct metric measurements, which are particularly useful in manipulation scenarios requiring accurate distances and object dimensions.


Semantic understanding can also be performed directly in 3D. Methods such as RandLA-Net~\cite{randlanet2020} and the Point Transformer family~\cite{wu2024ppt,pointcept2023} 
 provide powerful tools for point-cloud semantic segmentation. However, while they achieve high segmentation performance for the target data distribution, their vocabulary and generalization remain limited. Our approach targets open-vocabulary understanding without scene-specific training.


Foundation segmentation and vision-language models offer an alternative. SAM and SAM~2 enable class-agnostic image segmentation, separating the problem of locating scene entities from assigning them predefined semantic classes~\cite{kirillov2023segany,ravi2024sam2}. SAM~3 extends this paradigm by detecting, segmenting, and tracking all instances matching a text or a visual concept prompt~\cite{carion2025sam3segmentconcepts}. Region-conditioned multimodal models such as Describe Anything~\cite{lian2025describe} can generate detailed descriptions conditioned on selected image regions, while other approaches combine localized descriptions with spatial or geometric structure~\cite{huang2026pixels,Gorlo26cvpr-DAAAM}. Our work follows this direction but explicitly couples region-level semantics with metric RGB-D measurements for physical interaction. This decomposition is particularly attractive for robotics because mature RGB foundation models can be retained for semantic reasoning, while depth is processed independently to recover metric geometry.

\subsection{Open-Vocabulary 3D Representations}


Recent works combine 2D foundation model semantics with explicit 3D representations. ConceptFusion~\cite{jatavallabhula2023conceptfusionopensetmultimodal3d} fuses open-set visual features with geometric reconstruction to construct queryable multimodal 3D maps. ConceptGraphs~\cite{gu2024conceptgraphs} builds object-centric scene graphs with open-vocabulary features and spatial relations. OpenMask3D~\cite{takmaz2023openmask3d} and Open3DIS~\cite{nguyen2023open3dis} similarly combine 3D geometry with masks or features obtained from 2D foundation models to perform open-vocabulary 3D instance segmentation.


These approaches demonstrate the effectiveness of persistent 3D mapping, retrieval, and open-vocabulary segmentation of large scenes. In contrast, we focus on the complementary problem of extracting from a {single RGB-D observation} a compact object-centric representation for downstream robotic reasoning, associating each entity with natural-language semantics and directly measured metric information.


\subsection{Structured Representations for Task Planning}


Explicit scene structure becomes particularly important in several robotic tasks, including assembly, where success depends not only on recognizing components but also on recovering attributes, poses, dimensions, and spatial relationships. Neural Assembler~\cite{yan2024neuralassemblerlearninggenerate} estimates component masks, poses, relationships, and assembly sequences from multi-view observations, WorkBenchMark~\cite{ma2026workbenchmarklegobasedassemblybenchmark} studies LEGO Duplo assembly through structured reasoning over object properties and relative poses. These works highlight the value of object-centric representations, but focus on task-specific assembly settings.


In contrast, our work separates {scene perception from task reasoning}. From a single RGB-D observation, we construct a generic object-centric representation without object-specific perception models or scene-specific fine-tuning. The resulting knowledge base can subsequently be interpreted according to the task at hand, and interfaced with higher-level reasoning and planning approaches~\cite{saccon2026combininglargelanguagemodels}. This separation is intended to preserve the generality of modern foundation models while exposing the explicit metric structure required for robotic manipulation.

\section{Methodology}\label{sec:methodology}

\subsection{Scene Understanding Pipeline}
\label{ssec:SceneUnderstandingPipeline}

We consider the problem of constructing a task-independent, object-centric
representation of a previously unseen workspace from a single RGB-D
observation
\begin{equation}
    \cameraimage =
    \left\langle I^\mathrm{RGB}, I^\mathrm{DEPTH} \right\rangle ,
\end{equation}
where $I^\mathrm{RGB}\in\mathbb{R}^{H\times W\times 3}$ and
$I^\mathrm{DEPTH}\in\mathbb{R}^{H\times W}$ denote the RGB and metric depth images,
respectively.
Our framework explicitly decouples semantic interpretation from metric
geometric estimation. The RGB observation is first decomposed into candidate
scene entities. Each retained entity is semantically described by a
VLM, while the same image-space mask is
used to associate the entity with the corresponding depth measurements.
The two branches are then combined into a unified object-centric
representation for downstream reasoning and robotic tasks (Fig.~\ref{fig:diagram}).
Importantly, this representation is constructed independently of the task
subsequently assigned to the robot, allowing the same perceived scene to
support different natural-language instructions.

\subsection{Scene Decomposition}
\label{ssec:SceneDecomposition}

Given $I^\mathrm{RGB}$, a class-agnostic segmentation model $\mathcal{S}$ produces
an initial set of binary masks
\begin{equation}
    \mathcal{M}' =
    \left\{ M_j \right\}_{j=1}^{N'},
    \quad
    M_j\in\{0,1\}^{H\times W}.
\end{equation}
The initial proposals are filtered to remove unsuitable regions, including
small or redundant segments, background regions, and dominant planar
structures such as floors, walls, or supporting surfaces.
Sec.~\ref{sec:experiments} provides more details on this matter. 
The resulting set is
\begin{equation}
    \mathcal{M}
    =
    \left\{(M_j,b_j)\right\}_{j=1}^{N},
    \quad N\leq N',
\end{equation}
where $b_j$ denotes the bounding box associated with $M_j$.

For each retained segment, an object-focused observation $I^\mathrm{obj}_j$ is
extracted by applying the binary mask to the RGB image,

\begin{equation}
    I^\mathrm{obj}_j =
    \operatorname{Mask}\!\left(I^\mathrm{RGB},M_j\right).
\end{equation}
The segmentation mask therefore provides the common image-space reference
used to associate semantic and geometric information with the same physical
entity, and hence we will prefer using $M_j$ to $I^\mathrm{obj}_j$ afterwards.

\subsection{Context-Aware Semantic Annotation}
\label{ssec:SemanticAnnotation}

Each retained entity $I^\mathrm{obj}_j$ is independently annotated by a VLM. However, using only the
isolated object region $I^\mathrm{obj}_j$ may remove contextual information useful for
disambiguating its identity or function, whereas querying the complete image $I^\mathrm{RGB}$
alone does not explicitly identify the entity to be described. We therefore
condition the annotation on both the complete scene and the localized object,
\begin{equation}
    s_j =
    \mathcal{A}\left(
        I^\mathrm{RGB},
        I^\mathrm{obj}_j
    \right),
\end{equation}
where $\mathcal{A}$ denotes the multimodal annotation model and $s_j$ the
resulting semantic description.
The complete image provides scene context, while $I^{obj}_j$ identifies the
specific entity of interest. The resulting description $s_j$ may include its
identity, visual attributes, functional properties, affordances, and
qualitative relationships with surrounding elements. Metric quantities are
instead recovered independently from the depth observation.

\subsection{Geometric-Semantic Association}
\label{ssec:MetricGeometricGrounding}

Let $D\in\mathbb{R}^{H\times W}$ denote the metric depth map aligned with the
RGB observation. For each object mask, we first collect the image coordinates
with valid depth measurements,
\begin{equation}
    \Omega_j =
    \left\{
        (u,v)
        \;\middle|\;
        M_j(u,v)=1,\;
        D(u,v)>0,\;
        D(u,v)\text{ finite}
    \right\}.
\end{equation}
We define the representative object depth $d_j$ as the median of the valid depth
measurements contained within the mask,
\begin{equation}
    d_j =\operatorname{median}_{(u,v)\in\Omega_j} D(u,v).
\end{equation}
The median provides a robust estimate of the object range, reducing the
influence of isolated erroneous measurements and pixels incorrectly assigned
to the object near segmentation boundaries.

The object mask $M_j$ provides the image-space localization of the object.
We denote its centroid as
\begin{equation}
    {p}_j =
    \left(c^x_j,c^y_j\right).
\end{equation}
Together with the representative metric depth $d_j$, this defines the
geometric information associated with object $j$ as
\begin{equation}
    g_j =
    \left\langle
        {p}_j,\,
        d_j
    \right\rangle .
\end{equation}


%
It is worth mentioning that the same formulation can also accommodate metric monocular depth estimation
when measured depth is unavailable, as investigated in
Sec.~\ref{sec:experiments}.

\newcommand{\figsize}{0.195}
\begin{figure*}[t]
    \centering
    \includegraphics[width=\figsize\textwidth]{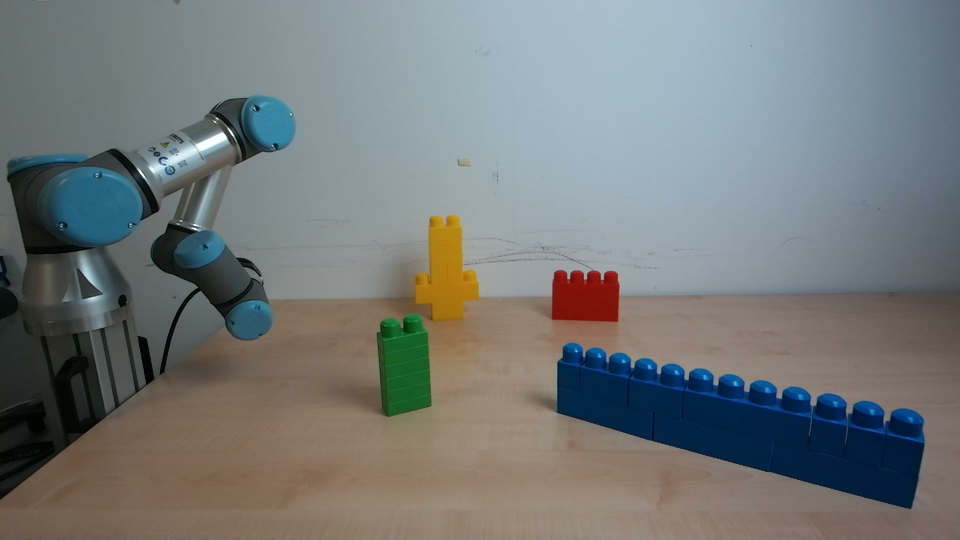}
    \includegraphics[width=\figsize\textwidth]{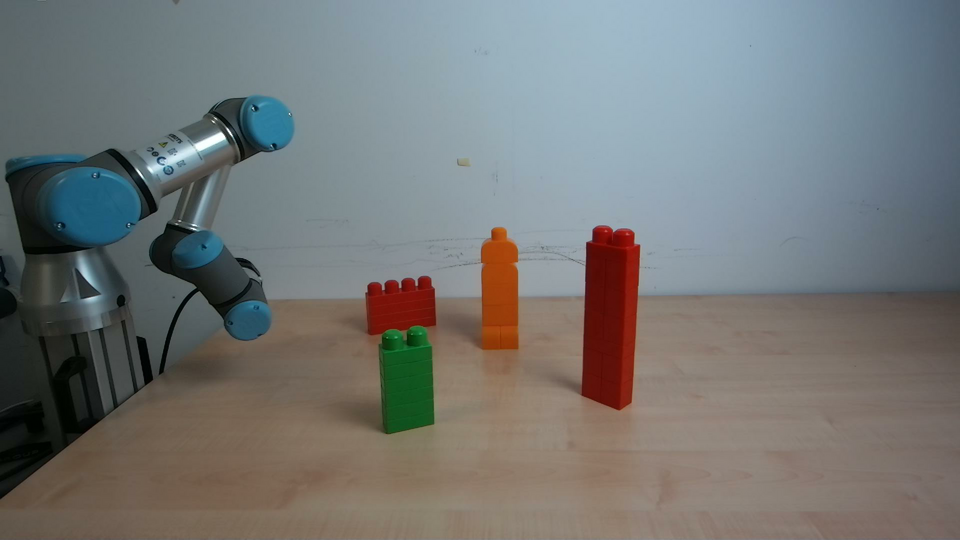}
    \includegraphics[width=\figsize\textwidth]{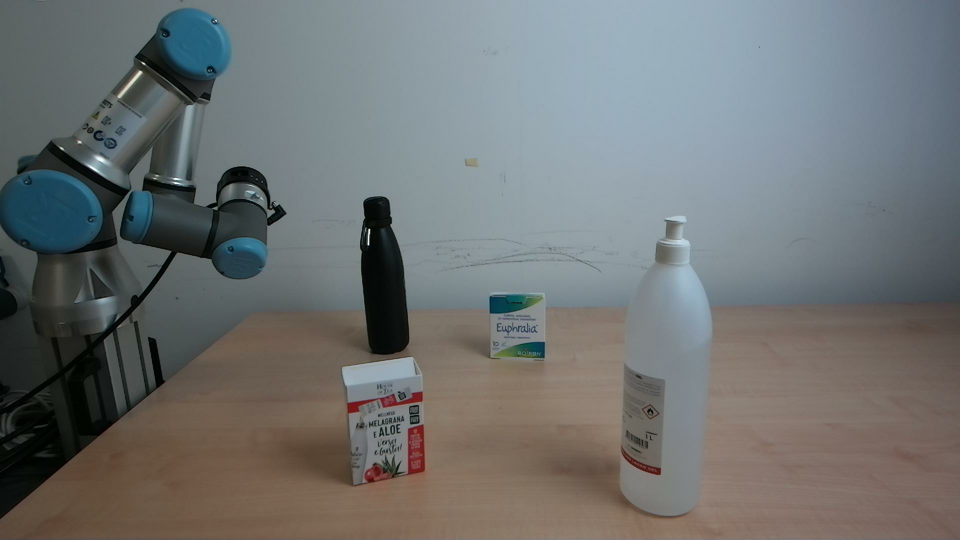}
    \includegraphics[width=\figsize\textwidth]{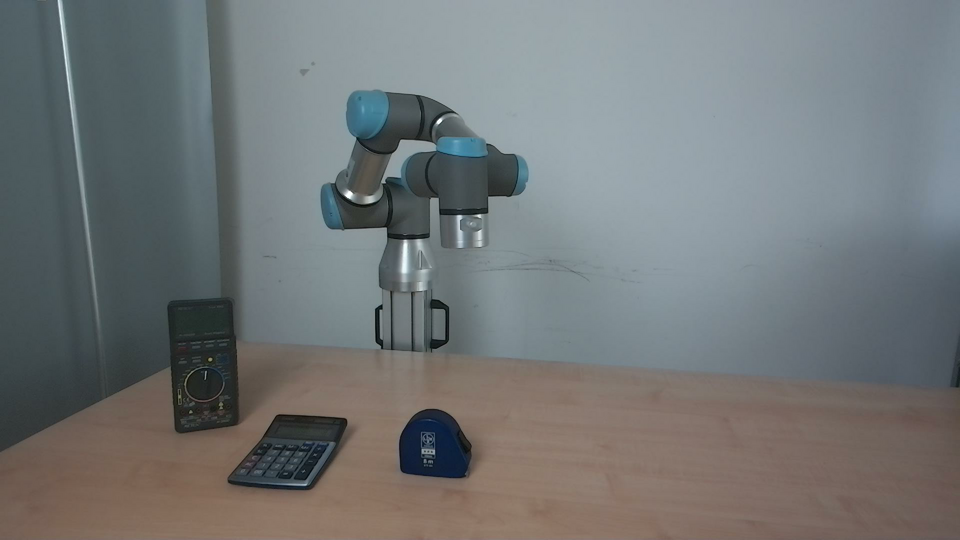}
    \includegraphics[width=\figsize\textwidth]{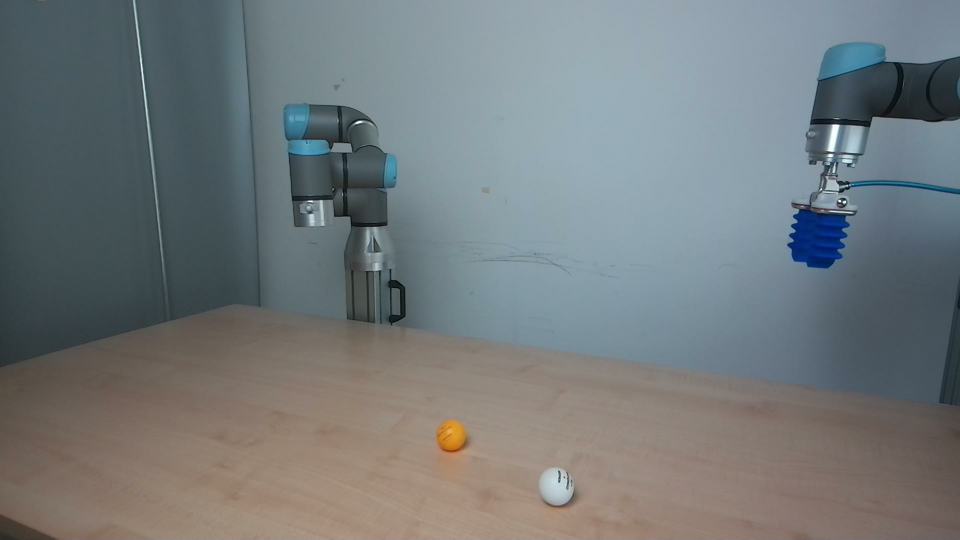} \\ 
    \includegraphics[width=\figsize\textwidth]{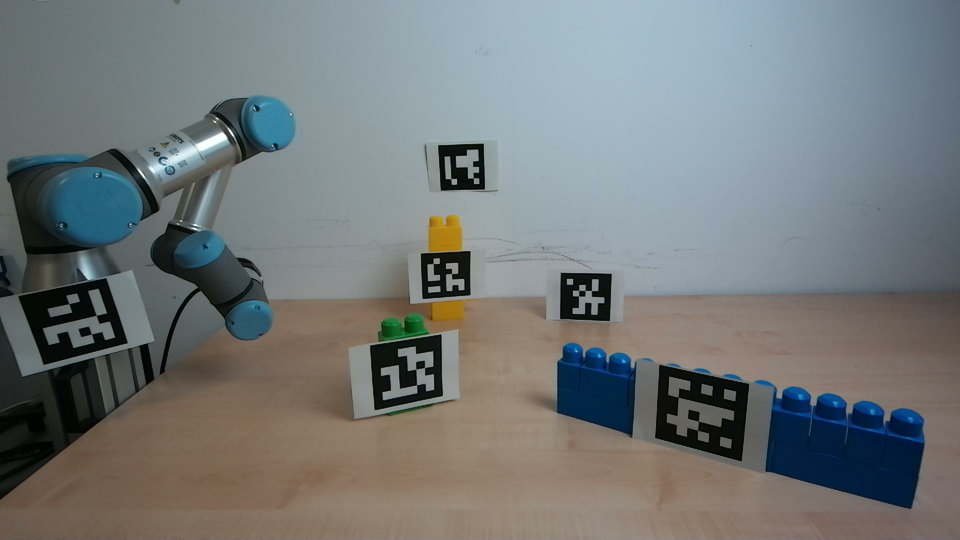}
    \includegraphics[width=\figsize\textwidth]{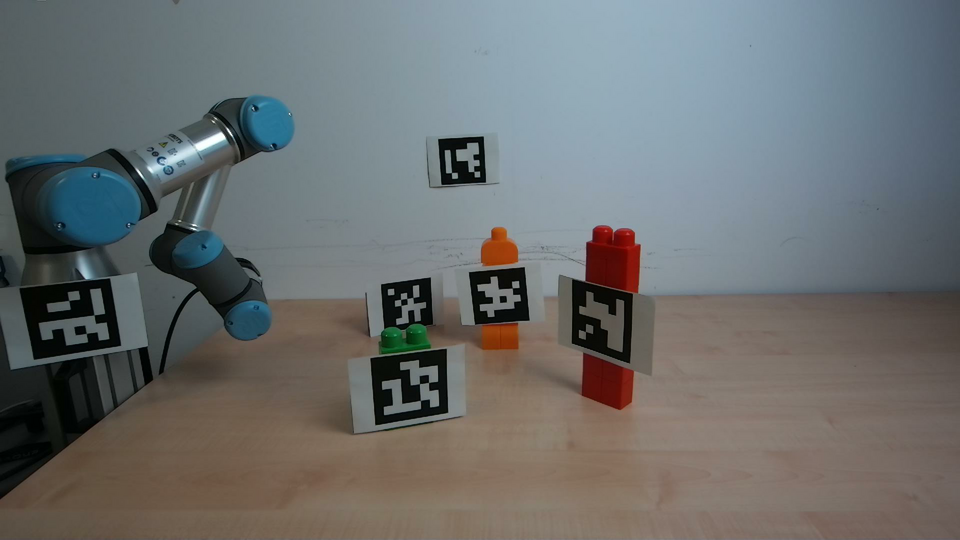}
    \includegraphics[width=\figsize\textwidth]{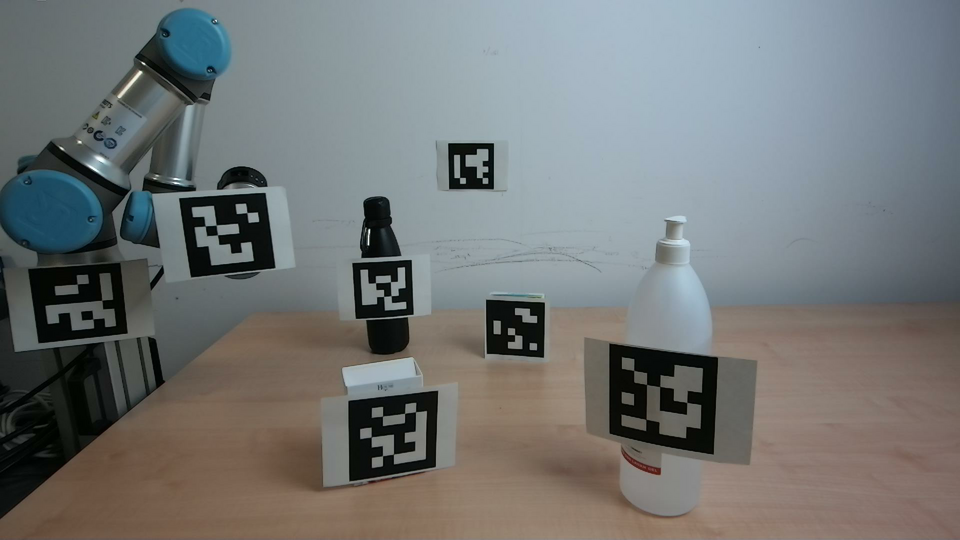}
    \includegraphics[width=\figsize\textwidth]{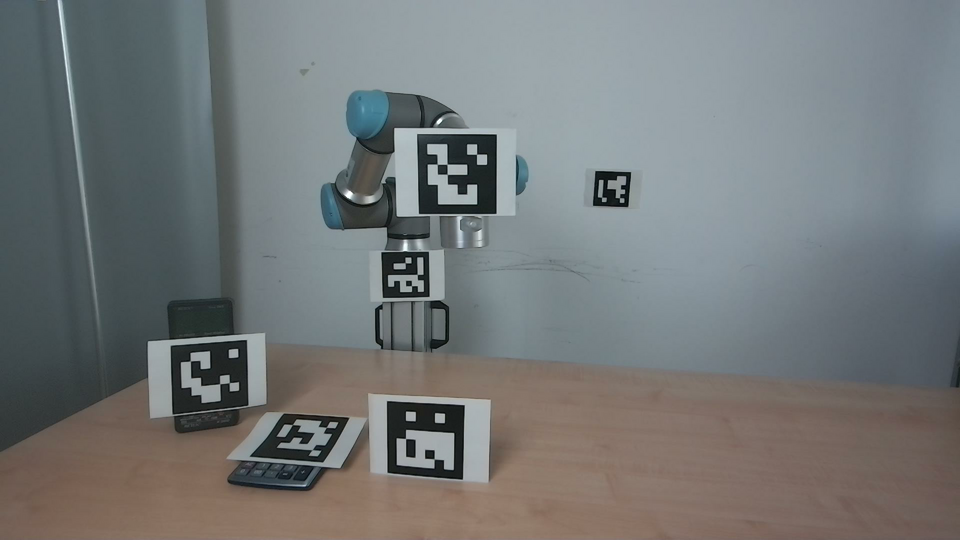}
    \includegraphics[width=\figsize\textwidth]{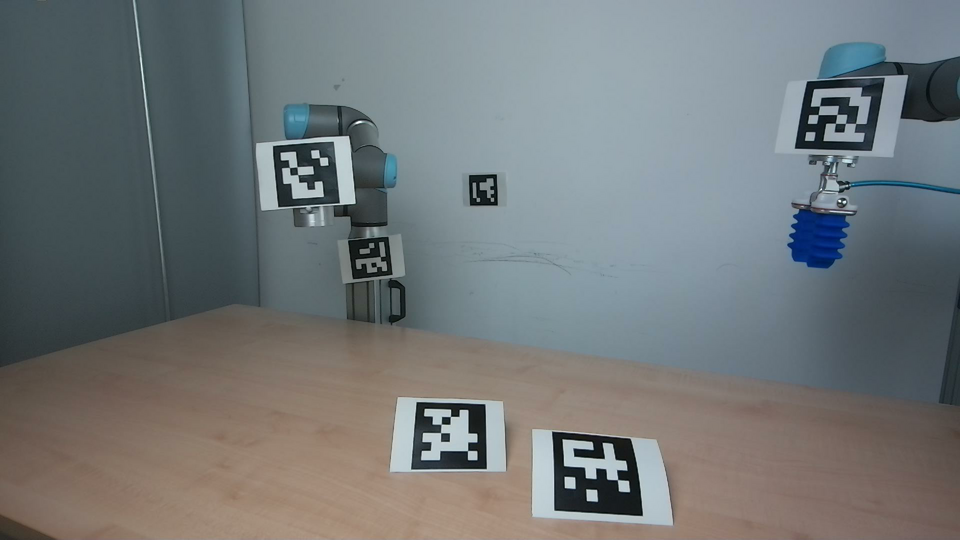} \\
    \includegraphics[width=\figsize\textwidth]{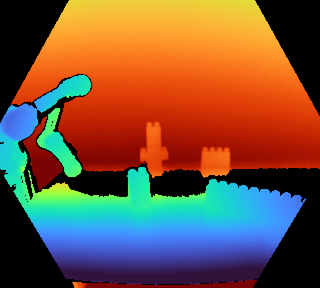}
    \includegraphics[width=\figsize\textwidth]{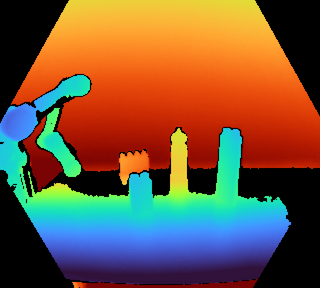}
    \includegraphics[width=\figsize\textwidth]{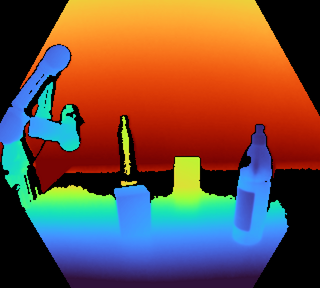}
    \includegraphics[width=\figsize\textwidth]{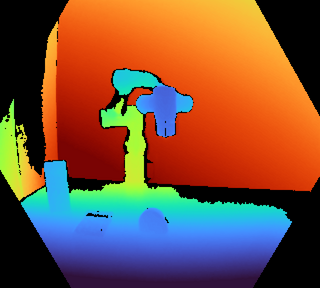}
    \includegraphics[width=\figsize\textwidth]{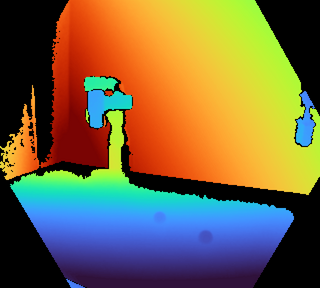} \\
    \caption{Representative images from the collected dataset. Top row: clean RGB observations. Middle row: corresponding ArUco augmented images. Bottom row: associated depth observations, captured with the Orbbec Femto Mega camera.}
    \label{fig:dataset}
\end{figure*}

\subsection{Object-Centric Scene Representation}
\label{ssec:ObjectCentricSceneRepresentation}
Semantic and geometric information are finally combined into an
object-level representation
\begin{equation}
    o_j = \left\langle M_j, s_j, g_j \right\rangle,
\end{equation}
and the complete scene is represented as
\begin{equation}
    \mathcal{O} = \{o_j\}_{j=1}^{N}.
\end{equation}
The segmentation mask $M_j$ acts internally as the common image-space
reference used to associate the semantic and geometric branches, while
the masked observation $I_j^{obj}$ is used only as visual input to the
annotation model.

Each semantic description is therefore explicitly grounded to the
corresponding physical region of the workspace. The resulting representation
can be interpreted both as a semantically annotated geometric reconstruction
and as a compact object-centric knowledge base.
Indeed, given a natural-language instruction, a downstream reasoning module can query
$\mathcal{O}$ to identify relevant entities, reason over their semantic properties and relationships, and retrieve their metric information for
planning or manipulation. Since $\mathcal{O}$ is constructed independently of
the task, the same scene representation can be reused across different
instructions.
Human instructions can therefore be interpreted with respect to the reconstructed scene, allowing the system to identify relevant entities and spatial relations and to derive the scene information required for downstream robotic tasks.
The framework is also modular with respect to the segmentation, annotation,
and depth-estimation backends, which can be replaced independently while
preserving the same object-level interface.


\section{Dataset}\label{sec:dataset}

To evaluate the proposed framework, we collected an RGB-D dataset comprising 151 tabletop scene configurations using an Orbbec Femto Mega camera. For each scene, the dataset contains: \textit{1)} an RGB image with a resolution of $1920\times1080$ pixels and \textit{2)} a corresponding $640\times576$ 16-bit depth image. Since the RGB and depth sensors have different formats, the raw depth observations are not directly pixel-aligned with the RGB images, but are registered as described in Sec.~\ref{ssec:MetricGeometricGrounding}. Examples of scenes in the dataset are shown in Fig.~\ref{fig:dataset}.

The scenes were constructed by varying the number, identity, and spatial arrangement of the objects placed within the robot workspace. Overall, the dataset contains 26 distinct objects, including blocks with different shapes and colors, bottles, boxes, tools, measurement devices, a cup, a wallet, and ping-pong balls. Each scene contains between one and five task objects. The robot base and end-effector are additionally represented in the reference annotations when visible. This results in scenes with objects presenting substantially different appearance, size, shape, and semantic properties, while repeated occurrences of the same objects at different locations allow the consistency of the perception pipeline to be evaluated across different scene configurations.

For quantitative evaluation, each clean observation is associated with a second RGB image in which ArUco markers are introduced as pose and semantic references. A dedicated marker is associated with each object identity, while an additional marker defines the world reference frame. Given the calibrated camera parameters and the known marker size, the detected fiducials provide reference object identities and their geometric position with respect to the camera and world frames. Importantly, the marker-augmented images are used only to construct the reference annotations and are never provided as input to the proposed perception framework. All segmentation and semantic reasoning experiments operate exclusively on the corresponding clean RGB images.
Moreover, the dataset is used exclusively for evaluation: none of the segmentation, vision-language, or depth-estimation models are trained or fine-tuned on these images. This setup is consistent with the objective of evaluating the framework under a zero-shot, task-independent perception setting.

\section{Experimental Evaluation}\label{sec:experiments}
\begin{table*}[t]
    \centering
    \caption{The results obtained when using a VLM to do annotation and geometric grounding. The model was in charge of annotating the objects in the environment, producing the position in the image of the object and estimating the depth from the camera.}
    \label{tab:onlyvlm}
    \begin{tabular}{lccccc}
    \toprule
    & {\sc Semantic Annotation} & \multicolumn{2}{c}{{\sc Position Error} [px]} & \multicolumn{2}{c}{{\sc Depth Error} [mm]} \\
    {\sc Models} & {\sc Recall} & {\sc Median} & {\sc RMSE} & {\sc Median} & {\sc RMSE} \\
    \midrule
    Sonnet 4.6   &         92.8\% &           29.3 &           70.9 &           96.7 &          166.4 \\
    GPT 5.4      &         95.9\% &           16.8 &           48.6 &          151.6 &          241.0 \\
    GPT 5.4-Mini &         90.0\% &           20.4 &           48.2 &          138.9 &          216.3 \\
    GPT 5.4-Nano &         60.3\% &          124.6 &          199.4 &          158.5 &          260.6 \\
    Qwen 3-VL    &         79.2\% &          221.7 &          238.1 &          413.9 &          519.5 \\
    Qwen 3.6     &         92.4\% &          240.5 &          252.4 &          192.3 &          228.4 \\
    \bottomrule
    \end{tabular}
\end{table*}


The experiments clarify the central claim of the proposed framework: while VLMs provide strong semantic understanding of visual scenes, metric geometric information is more reliably obtained by specialized perception tools. We therefore compare direct VLM inference, in which both semantic annotations and object depth are inferred from the RGB image, against the proposed architecture, where the VLM is used for semantic reasoning while geometric information is obtained independently and associated with the corresponding segmented objects.

\subsection{Evaluation Protocol and Metrics}
For semantic evaluation, predictions are associated with the reference annotations through their semantic tags. For each scene, predicted and ground-truth tag names are normalized and semantic recall is computed as the fraction of ground-truth tags appearing among the predictions. The recall reported in the following tables is the average across all dataset scenes.

For geometric evaluation, a predicted object is associated with the ArUco reference carrying the same normalized semantic tag. Image-space localization error is computed as the Euclidean distance between the center of the predicted bounding box and the reference position obtained from the corresponding ArUco marker, and is reported through median error and root mean squared error (RMSE) in pixels. Depth is evaluated against the metric distance obtained from the calibrated ArUco reference. We report the median absolute error and RMSE in millimeters, considering only matched objects for which a valid depth estimate is available.

\subsection{VLM-Only Baseline}\label{ssec:vlmBaseline}
This experiment aims at showing that VLMs alone are not sufficient for accurate 3D scene interpretation. To this end, we consider the zero-shot use of the VLM as the baseline for this work. For this first experiment, we use a pool of available VLMs, considering both proprietary and open-source models. As proprietary models, we considered two alternatives from Anthropic and OpenAI. From Anthropic, we used Claude Sonnet 4.6, setting only the maximum number of tokens used to $8192$, sufficiently above the expected structured output length to avoid truncation. From OpenAI, we tested three models, namely GPT 5.4, GPT 5.4-Mini and GPT 5.4-Nano, to show the differences that models of the same family can have. For these models, we set the max token usage to $16384$ and the seed to $42$. Both Sonnet and the GPT models were queried through the Azure SDK. We also included two models from Qwen that are open-source and that can be run locally. The two models that were considered are Qwen3-VL-4B-Instruct\footnote{\label{fn:qwen3-vl}\url{https://huggingface.co/Qwen/Qwen3-VL-4B-Instruct}} and Qwen3.6-35B-A3B\footnote{\label{fn:qwen36}\url{https://huggingface.co/Qwen/Qwen3.6-35B-A3B}}, later referred to as Qwen 3-VL and Qwen 3.6, respectively. These models were run on a server equipped with one Nvidia A100 with 80GB of VRAM. As for the parameters, for both models, we set the maximum number of generated tokens to $8192$, the temperature to $0$, \texttt{top\_p} (nucleus sampling) to $1.0$, and the random seed to $42$. The combination of zero temperature and $\texttt{top\_p} = 1.0$ results in deterministic greedy decoding. 

For each model, the prompt provided a single RGB image together with the set of admissible semantic labels and asking the model, in an open-vocabulary setting, to jointly predict the object classes, their image-space bounding boxes, and their metric distance from the camera. The models were not fine-tuned for this particular task, nor were they provided with in-context demonstrations. The results in Table~\ref{tab:onlyvlm} show that the models are reliable at identifying objects and providing the correct tags for them, with GPT 5.4 reaching a $95.9$\% recall. GPT 5.4 and 5.4-mini, as well as Sonnet 4.6, are also good at estimating the 2D position of the objects in the scene, with median errors comparable to the ones of the whole pipeline shown in Table~\ref{tab:pipelineResults}. However, the models fall short when having to estimate the depth of the objects in the scene.  This implies an inaccurate estimation of the 3D position, which is not compatible with typical robot manipulation tasks. 


\begin{table*}[t]
    \centering
    \caption{Experiments run for the pipeline. For each combination of segmentation model, annotation model and depth tool, we report semantic recall, the median and root mean squared error (RMSE) of the image-space position in pixels, and the median absolute error and RMSE of the depth estimate in millimeters. }
    \label{tab:pipelineResults}
    \begin{tabular}{llccccc}
    \toprule
    \multicolumn{2}{c}{{\sc Tools}} & {\sc Semantic Annotation} & \multicolumn{2}{c}{{\sc Position Error}} & \multicolumn{2}{c}{{\sc Depth Error}} \\
    {\sc Segmentation} & {\sc Annotation} & {\sc Recall [\%]} & {\sc Median} & {\sc RMSE} & {\sc Median} & {\sc RMSE} \\
    \midrule
    SAM3         & Sonnet 4.6   &           90.3 &           16.1 &          164.0 &           11.4 &           91.0 \\
    SAM3         & GPT 5.4      &           94.1 &           15.2 &           77.9 &           11.6 &           84.3 \\
    SAM3         & GPT 5.4-Mini &           88.0 &           16.2 &          149.9 &           11.9 &           89.6 \\
    SAM3         & GPT 5.4-Nano &           61.3 &           21.0 &          216.4 &           13.8 &          104.3 \\
    SAM3         & Qwen 3-VL    &           76.2 &           19.6 &          128.1 &           12.2 &          118.9 \\
    SAM3         & Qwen 3.6     &           91.7 &           14.8 &           74.7 &           11.6 &           92.7 \\
    SAM2-l       & GPT 5.4      &           93.5 &           13.6 &           65.3 &           17.5 &           89.0 \\
    SAM1-h       & GPT 5.4      &           93.6 &           14.1 &           67.3 &           13.2 &          108.7 \\
    FastSAM      & GPT 5.4      &           92.8 &           20.4 &          159.5 &           15.4 &          175.8 \\
    \bottomrule
    \end{tabular}
\end{table*}
\subsection{Pipeline experiments}\label{ssec:pipelineExperiments}
For the pipeline, we chose different models both for the segmentation and the annotation. For the segmentation, we considered SAM3~\cite{carion2025sam3segmentconcepts}, SAM2~\cite{ravi2024sam2} (with checkpoint \texttt{sam2-l}), SAM1~\cite{kirillov2023segany} (with checkpoint \texttt{sam1-h}), and FastSAM~\cite{zhao2023fastsegment}. For the annotation, we considered the same models tested in Sec.~\ref{ssec:vlmBaseline}. For SAM3, we report the results in combination with all the VLM models, while for SAM1, SAM2, and FastSAM, we report only the best results (obtained with GPT 5.4). 

As shown in Fig.~\ref{fig:diagram}, we used a VLM to precondition the SAM3 model. In particular, we used the same VLM employed for the subsequent semantic annotation stage, so as not to introduce an additional model-dependent variable in the comparison. The preconditioning model was asked to identify all visible non-background objects in the scene and represent them as short noun phrases, using visual attributes such as color, size, and overall shape to distinguish different object types. The prompt additionally included a small set of in-context examples illustrating suitable and unsuitable concept formulations, and the resulting phrases were then provided to SAM3 as segmentation prompts.

For SAM1, SAM2, and FastSAM, we adopted a different approach: we relied on their automatic and class-agnostic segmentation capabilities, without semantic preconditioning. The models were first used to generate candidate masks over the entire image, first removing background regions and redundant or too small masks. The remaining object masks were then passed independently to the VLM for semantic annotation.

The VLMs used for the semantic annotation, instead, were instructed to assign to each segmented object one of the predefined semantic tags, using both the full scene and the corresponding object crop as visual context, without providing task-specific examples or demonstrations.

The following experiments were carried out exploiting the depth metadata from the RGB-D images. The error on the position is computed w.r.t the bounding box of the masks, and the depth is extracted as the median of the depth values of the pixels in the mask.  

The results in Table~II show that introducing specialized perception components substantially improves the geometric information associated with each object, while preserving a semantic recall comparable to the VLM-only baseline. For instance, SAM3 with GPT~5.4 achieves a recall of 94.1\%, compared with 95.9\% for the corresponding VLM-only baseline. The differences in depth estimation are particularly pronounced. Across the six VLM backends evaluated with SAM3, the average per-configuration median depth error decreases from 192.0~mm with direct VLM inference to 12.1~mm with the proposed pipeline, corresponding to an absolute reduction of 179.9~mm, or 93.7\%.


Overall, experiments in Sec.~\ref{ssec:vlmBaseline}~and~\ref{ssec:pipelineExperiments} support the main hypothesis of this work: VLMs are effective semantic annotators, but they are less reliable when having to infer geometric quantities. Enhancing the scene by leveraging simple perception tools for localization and depth estimation yields a more accurate representation without sacrificing the semantic accuracy of VLMs.


\subsection{Depth Techniques Ablation}\label{ssec:ablation}
For the results reported in Sec.~\ref{ssec:pipelineExperiments}, we used measured depth data from the RGB-D pictures, and computed the depth as the median of the pixels in the mask. We also carried out experiments using a depth prediction model, namely Depth Anything V3~\cite{depthanything3} (DA3). This is useful to investigate the possibility of deploying the proposed pipeline when only RGB images are available. Moreover, we also tested a na{\"i}ve technique to compute the depth of the objects by taking the depth of the pixel at the center of the bounding box of the masks. This ablation evaluates whether exploiting the complete object region provides an advantage over a single representative image location. 

\begin{table*}
    \centering
    \caption{Depth estimation ablation for RGB-D and Depth Anything 3 using mask-median and bounding-box-center depth extraction. Median absolute error and RMSE are reported in millimeters.}
    \label{tab:depthAblationStudy}
    \begin{tabular}{llcccccccc}
    \toprule
    \multicolumn{2}{c}{{\sc Tools}} & \multicolumn{4}{c}{{\sc RGB-D}} & \multicolumn{4}{c}{{\sc DA3}} \\
    \cmidrule(lr){3-6} \cmidrule(lr){7-10}
    & & \multicolumn{2}{c}{{\sc Mask median}} & \multicolumn{2}{c}{{\sc BBox center}} & \multicolumn{2}{c}{{\sc Mask median}} & \multicolumn{2}{c}{{\sc BBox center}} \\
    \cmidrule(lr){3-4} \cmidrule(lr){5-6} \cmidrule(lr){7-8} \cmidrule(lr){9-10}
    {\sc Segmentation} & {\sc Annotation} & {\sc Median} & {\sc RMSE} & {\sc Median} & {\sc RMSE} & {\sc Median} & {\sc RMSE} & {\sc Median} & {\sc RMSE} \\
    \midrule
    SAM3         & Sonnet 4.6   &           11.4 &           91.0 &           14.5 &          215.4 &           70.9 &          132.3 &          105.3 &          186.2 \\
    SAM3         & GPT 5.4      &           11.6 &           84.3 &           14.4 &          209.2 &           69.8 &          127.5 &           98.7 &          181.2 \\
    SAM3         & GPT 5.4-Mini &           11.9 &           89.6 &           15.3 &          216.5 &           68.8 &          127.1 &           99.2 &          183.9 \\
    SAM3         & GPT 5.4-Nano &           13.8 &          104.3 &           15.7 &          187.6 &           77.5 &          131.9 &          102.0 &          174.7 \\
    SAM3         & Qwen 3-VL    &           12.2 &          118.9 &           19.4 &          241.6 &           70.5 &          141.1 &          116.7 &          199.1 \\
    SAM3         & Qwen 3.6     &           11.6 &           92.7 &           14.4 &          214.9 &           70.5 &          128.9 &          100.2 &          183.7 \\
    SAM2-l       & GPT 5.4      &           17.5 &           89.0 &           15.5 &          163.4 &           67.1 &          120.6 &           88.7 &          158.6 \\
    SAM1-h       & GPT 5.4      &           13.2 &          108.7 &           14.9 &          217.0 &           66.6 &          135.5 &           97.4 &          174.3 \\
    FastSAM      & GPT 5.4      &           15.4 &          175.8 &           18.3 &          237.3 &           67.7 &          154.8 &           92.3 &          185.9 \\
    \bottomrule
    \end{tabular}
\end{table*}

\begin{figure*}[t]
    \centering
    \includegraphics[width=0.325\linewidth]{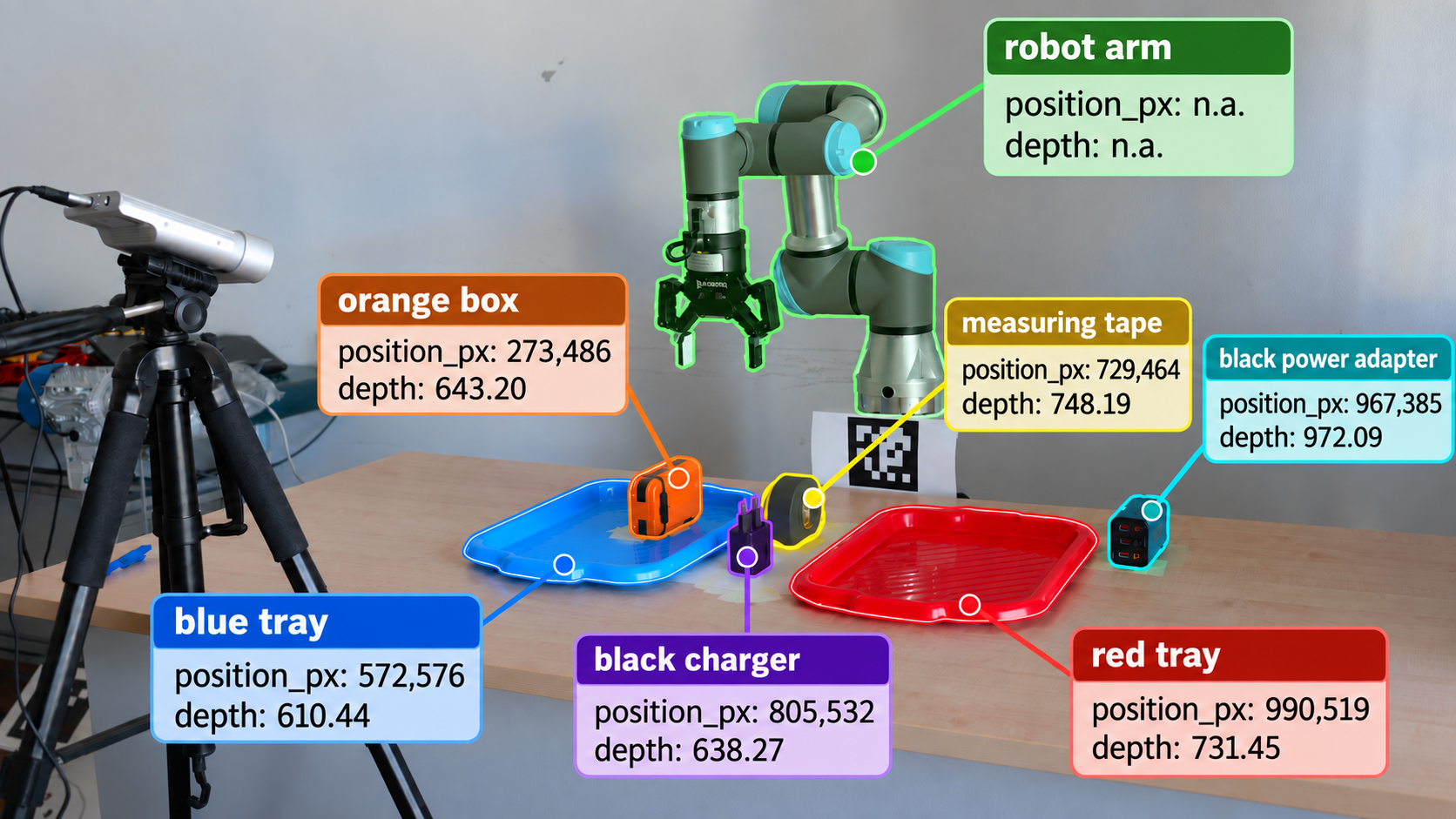}
    \includegraphics[width=0.325\linewidth]{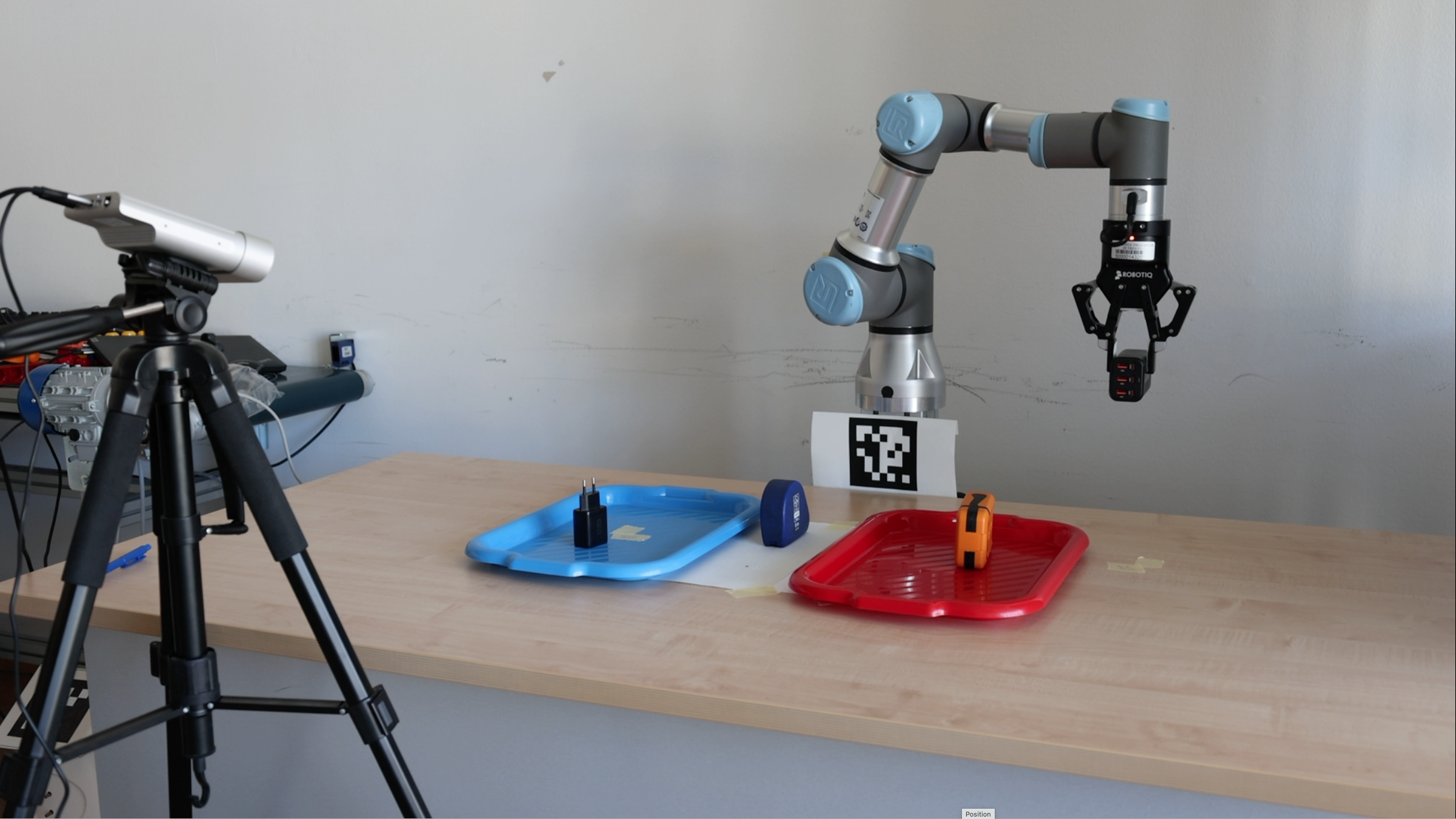}
    \includegraphics[width=0.325\linewidth]{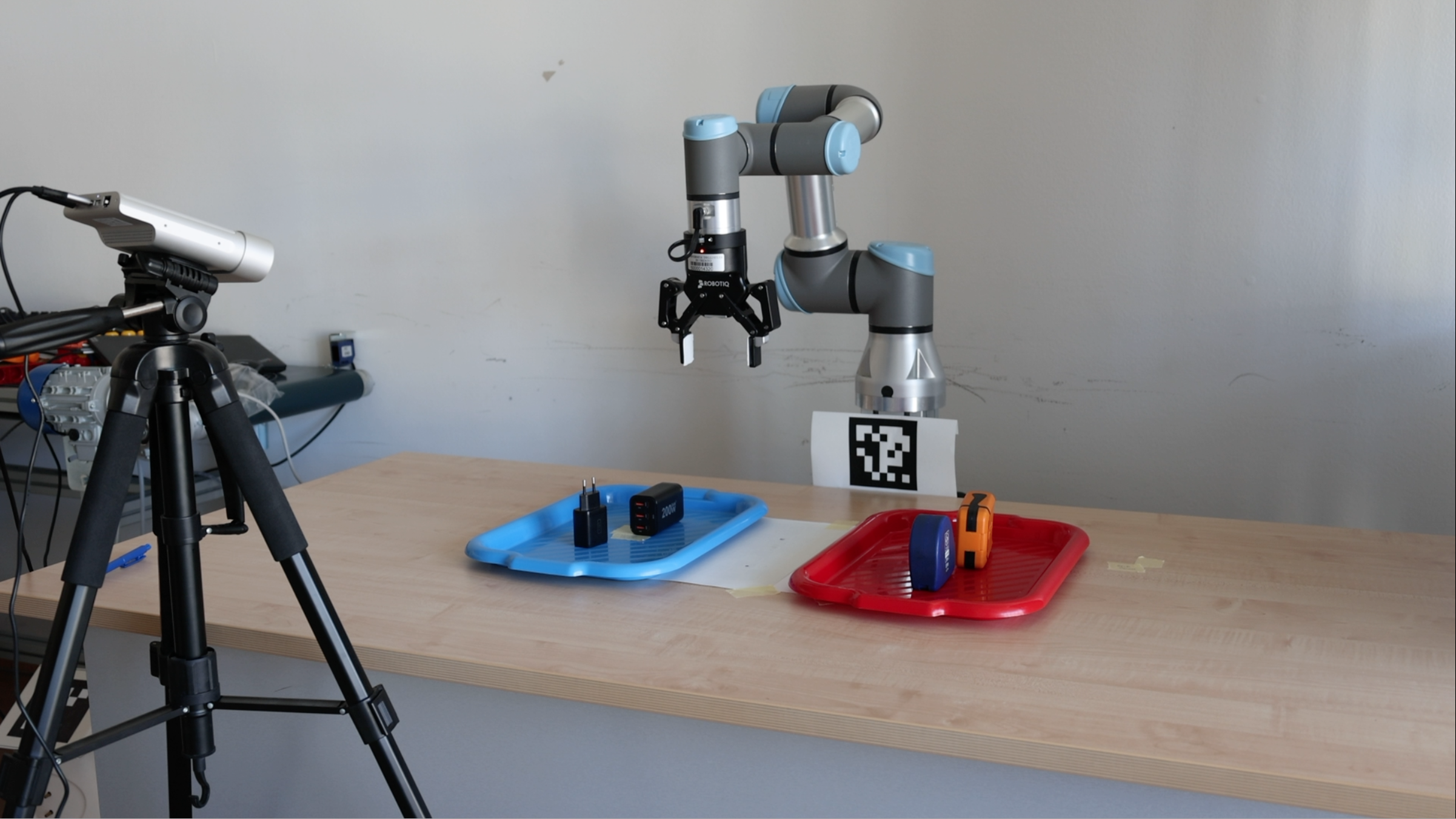}
\scriptsize
\begin{minted}[breaklines, breakindent=0pt, breaksymbolleft={}]{text}
In this scenario, a robotic arm must rearrange four objects located on a table containing a blue tray and a red tray. Initially, the orange box is inside the blue tray, while the measuring tape, black charger, and black power adapter are located on the table. The goal is to place the measuring tape and the orange box inside the red tray, and the black charger and the black power adapter inside the blue tray. The robotic arm can move an object from its current location to a target tray, with each move taking between 2 and 30 time units.
\end{minted}
    \caption{Above, three pictures of the real-world experiment showing the initial state (with overlays for the segmented objects), an intermediate configuration and the final state (from left to right). Below, the high-level task description passed to the task planner is shown.}
    \label{fig:realWorldExperiment}
\end{figure*}

The results of this study are shown in Table~\ref{tab:depthAblationStudy}. Focusing on the approach to compute the depth, it is clear that exploiting the complete segmentation mask provides a more reliable estimate than sampling a single pixel at the center of its bounding box. This trend is consistent across all segmentation and annotation combinations, both when using measured RGB-D data and when relying on DA3 predictions. Averaging the per-configuration median depth errors across all evaluated
configurations, the mask-median strategy reduces the median depth error from 15.8~mm to 13.2~mm with RGB-D data, and from 100.1~mm to 69.9~mm with DA3, corresponding to reductions of approximately 16.7\% and 30.1\%, respectively. The comparison also supports the use of the segmentation mask as the interface between the semantic and geometric components of the framework. Rather than relying on an arbitrary representative pixel, aggregating depth over the pixels belonging to the segmented object makes the estimate less sensitive to local prediction errors, missing depth measurements, and cases in which the center of the bounding box does not correspond to the visible surface of the target object.

Results also show that measured RGB-D information provides the most accurate geometric estimates. This is an expected result, since measured information is generally more accurate than an estimate. Nevertheless, the results obtained with DA3 remain substantially more accurate than the depth estimates produced directly by the VLMs (Table~\ref{tab:onlyvlm}). Across the same six VLM backends evaluated with SAM3, the average per-configuration median depth error decreases from 192.0~mm with direct VLM inference to 71.3~mm with DA3 and mask-median aggregation, corresponding to a reduction of 120.7~mm, or 62.8\%. Similarly, the average RMSE decreases from 272.0~mm to 131.5~mm, a reduction of 140.6~mm, or 51.7\%. This is particularly relevant when an RGB-D sensor is not available, as it shows that the proposed decomposition can still recover considerably more reliable metric information from a single RGB image by delegating depth estimation to a model specifically designed for geometric perception.


\subsection{Robot Demonstration}\label{ssec:realWorldDemonstration}
We carried out a robotic manipulation experiment (Fig.~\ref{fig:realWorldExperiment}) to showcase the complete workflow for executing a task from its natural language description and an RGB-D picture of the environment taken with an Orbbec Femto Mega camera. 

For the experiment, we considered a UR3 robot equipped with a Robotiq 2F-85 parallel gripper. The robot is asked to sort objects into two colored trays. As a proof of concept, we tested the combination proved to lead to the best results, as shown in Sec.~\ref{ssec:pipelineExperiments}. Hence, we used SAM3 with GPT 5.4, measured depth from the RGB-D camera, and mask-median depth extraction. We integrated the JSON file containing the annotations and the geometric data extracted by the pipeline into the framework of~\cite{saccon2026combininglargelanguagemodels}. This enabled the automatic generation of a behavior tree for the robot's actions, which was then executed through the ROS2 middleware. 

We conducted three tests with different objects and configurations. The first is depicted in Figure~\ref{fig:realWorldExperiment}. In a second experiment, we placed an orange block and a yellow block in the red tray and a red block in the space between the two trays. The goal was to place the yellow block in the blue tray, place the red block on top of it and finally place the orange block in between the two trays. Finally, we ran a test in which two blue blocks and two red blocks were scattered in the scenario and had to be positioned inside the corresponding blue and red trays.

\section{Conclusion}\label{sec:conclusion}
In this work, we presented a modular perception framework for constructing a task-independent object-centric representation of previously unseen robotic workspaces from a single RGB-D observation. The framework explicitly separates semantic interpretation from metric geometric estimation: foundation segmentation models identify candidate scene entities, VLMs provide their semantic annotations, and specialized geometric perception tools associate each entity with metric spatial information.

The framework addresses a current limitation of existing VLMs. Indeed, we empirically show that direct VLM inference provides strong semantic recognition, but considerably less reliable estimates of object localization and depth. By grounding the semantic predictions through segmentation masks and dedicated depth estimation, the proposed framework preserves comparable semantic performance while substantially reducing geometric error. The depth ablation further shows that aggregating depth over the segmented object region provides a more robust estimate than relying on a single bounding-box-center measurement, while monocular depth estimation offers a viable alternative when RGB-D sensing is unavailable. Finally, integration with the framework from~\cite{saccon2026combininglargelanguagemodels} demonstrates how the resulting representation can directly support downstream task planning and robotic execution.

Future work will investigate richer geometric representations, including object pose and complete shape estimation, and evaluate how perception errors propagate to task-level performance in more complex and less structured robotic environments.

\bibliographystyle{IEEEtran}
\bibliography{biblio}

@IEEEtranBSTCTL{BSTcontrol,
  CTLuse_forced_etal       = "yes",
  CTLmax_names_forced_etal = "3",
  CTLnames_show_etal       = "3"
}

@inproceedings{shah2023mutex,
	title        = {MUTEX: Learning Unified Policies from Multimodal Task Specifications},
	author       = {Rutav Shah and Roberto Mart{\'\i}n-Mart{\'\i}n and Yuke Zhu},
	year         = 2023,
	booktitle    = {7th Annual Conference on Robot Learning},
	url          = {https://openreview.net/forum?id=PwqiqaaEzJ}
}

@inproceedings{nasiriany2022sailor,
      title={Learning and Retrieval from Prior Data for Skill-based Imitation Learning},
      author={Soroush Nasiriany and Tian Gao and Ajay Mandlekar and Yuke Zhu},
      booktitle={Conference on Robot Learning (CoRL)},
      year={2022}
    }

@inproceedings{walke2023bridgedata,
    title={BridgeData V2: A Dataset for Robot Learning at Scale},
    author={Walke, Homer and Black, Kevin and Lee, Abraham and Kim, Moo Jin and Du, Max and Zheng, Chongyi and Zhao, Tony and Hansen-Estruch, Philippe and Vuong, Quan and He, Andre and Myers, Vivek and Fang, Kuan and Finn, Chelsea and Levine, Sergey},
    booktitle={Conference on Robot Learning (CoRL)},
    year={2023}
}

@misc{jatavallabhula2023conceptfusionopensetmultimodal3d,
      title={ConceptFusion: Open-set Multimodal 3D Mapping}, 
      author={Krishna Murthy Jatavallabhula and Alihusein Kuwajerwala and Qiao Gu and Mohd Omama and Tao Chen and Alaa Maalouf and Shuang Li and Ganesh Iyer and Soroush Saryazdi and Nikhil Keetha and Ayush Tewari and Joshua B. Tenenbaum and Celso Miguel de Melo and Madhava Krishna and Liam Paull and Florian Shkurti and Antonio Torralba},
      year={2023},
      eprint={2302.07241},
      archivePrefix={arXiv},
      primaryClass={cs.CV},
      url={https://arxiv.org/abs/2302.07241}, 
}

@inproceedings{gu2024conceptgraphs,
  title={Conceptgraphs: Open-vocabulary 3d scene graphs for perception and planning},
  author={Gu, Qiao and Kuwajerwala, Ali and Morin, Sacha and Jatavallabhula, Krishna Murthy and Sen, Bipasha and Agarwal, Aditya and Rivera, Corban and Paul, William and Ellis, Kirsty and Chellappa, Rama and others},
  booktitle={2024 IEEE International Conference on Robotics and Automation (ICRA)},
  pages={5021--5028},
  year={2024},
  organization={IEEE}
}

@inproceedings{takmaz2023openmask3d,
  title={{OpenMask3D: Open-Vocabulary 3D Instance Segmentation}},
  author={Takmaz, Ay{\c{c}}a and Fedele, Elisabetta and Sumner, Robert W. and Pollefeys, Marc and Tombari, Federico and Engelmann, Francis},
  booktitle={Advances in Neural Information Processing Systems (NeurIPS)},
year={2023}
}

@inproceedings{nguyen2023open3dis,
        title={Open3DIS: Open-Vocabulary 3D Instance Segmentation with 2D Mask Guidance}, 
        author={Phuc D. A. Nguyen and Tuan Duc Ngo and Evangelos Kalogerakis and Chuang Gan and Anh Tran and Cuong Pham and Khoi Nguyen},
        year={2024},
        booktitle={Proceedings of the IEEE/CVF Conference on Computer Vision and Pattern Recognition (CVPR)}
}

@article{lian2025describe,
    title={Describe Anything: Detailed Localized Image and Video Captioning}, 
    author={Long Lian and Yifan Ding and Yunhao Ge and Sifei Liu and Hanzi Mao and Boyi Li and Marco Pavone and Ming-Yu Liu and Trevor Darrell and Adam Yala and Yin Cui},
    journal={arXiv preprint arXiv:2504.16072},
    year={2025}
}

@inproceedings{huang2026pixels,
author = {Huang, Lihong and Zhong, Sheng-hua and Zhang, Zhi and Liu, Yan},
title = {From pixels to logic: a perception-reasoning decomposition framework for open-world referring expression comprehension},
year = {2026},
isbn = {978-1-57735-906-7},
publisher = {AAAI Press},
url = {https://doi.org/10.1609/aaai.v40i7.37419},
doi = {10.1609/aaai.v40i7.37419},
booktitle = {Proceedings of the Fortieth AAAI Conference on Artificial Intelligence and Thirty-Eighth Conference on Innovative Applications of Artificial Intelligence and Sixteenth Symposium on Educational Advances in Artificial Intelligence},
articleno = {563},
numpages = {9},
series = {AAAI'26/IAAI'26/EAAI'26}
}

@inproceedings{Gorlo26cvpr-DAAAM,
  title={Describe anything anywhere at any moment},
  author={Gorlo, Nicolas and Schmid, Lukas and Carlone, Luca},
  booktitle={Proceedings of the IEEE/CVF Conference on Computer Vision and Pattern Recognition},
  pages={35002--35013},
  year={2026}
}

@misc{shridhar2021cliportpathwaysroboticmanipulation,
      title={CLIPort: What and Where Pathways for Robotic Manipulation}, 
      author={Mohit Shridhar and Lucas Manuelli and Dieter Fox},
      year={2021},
      eprint={2109.12098},
      archivePrefix={arXiv},
      primaryClass={cs.RO},
      url={https://arxiv.org/abs/2109.12098}, 
}

@misc{huang2023voxposercomposable3dvalue,
      title={VoxPoser: Composable 3D Value Maps for Robotic Manipulation with Language Models}, 
      author={Wenlong Huang and Chen Wang and Ruohan Zhang and Yunzhu Li and Jiajun Wu and Li Fei-Fei},
      year={2023},
      eprint={2307.05973},
      archivePrefix={arXiv},
      primaryClass={cs.RO},
      url={https://arxiv.org/abs/2307.05973}, 
}

@misc{huang2024rekepspatiotemporalreasoningrelational,
      title={ReKep: Spatio-Temporal Reasoning of Relational Keypoint Constraints for Robotic Manipulation}, 
      author={Wenlong Huang and Chen Wang and Yunzhu Li and Ruohan Zhang and Li Fei-Fei},
      year={2024},
      eprint={2409.01652},
      archivePrefix={arXiv},
      primaryClass={cs.RO},
      url={https://arxiv.org/abs/2409.01652}, 
}

@misc{yan2024neuralassemblerlearninggenerate,
      title={Neural Assembler: Learning to Generate Fine-Grained Robotic Assembly Instructions from Multi-View Images}, 
      author={Hongyu Yan and Yadong Mu},
      year={2024},
      eprint={2404.16423},
      archivePrefix={arXiv},
      primaryClass={cs.CV},
      url={https://arxiv.org/abs/2404.16423}, 
}

@misc{ma2026workbenchmarklegobasedassemblybenchmark,
      title={WorkBenchMark: A LEGO-Based Assembly Benchmark with an Assembly-by-Disassembly Baseline for the Smart Manufacturing League}, 
      author={Wenbo Ma and Daniel Swoboda and Matteo Tschesche and Till Hofmann},
      year={2026},
      eprint={2606.19358},
      archivePrefix={arXiv},
      primaryClass={cs.RO},
      url={https://arxiv.org/abs/2606.19358}, 
}

@article{depthanything3,
  title={Depth Anything 3: Recovering the visual space from any views},
  author={Haotong Lin and Sili Chen and Jun Hao Liew and Donny Y. Chen and Zhenyu Li and Guang Shi and Jiashi Feng and Bingyi Kang},
  journal={arXiv preprint arXiv:2511.10647},
  year={2025}
}

@inproceedings{wang2025vggt,
  title={VGGT: Visual Geometry Grounded Transformer},
  author={Wang, Jianyuan and Chen, Minghao and Karaev, Nikita and Vedaldi, Andrea and Rupprecht, Christian and Novotny, David},
  booktitle={Proceedings of the IEEE/CVF Conference on Computer Vision and Pattern Recognition},
  year={2025}
}

@article{kirillov2023segany,
  title={Segment Anything},
  author={Kirillov, Alexander and Mintun, Eric and Ravi, Nikhila and Mao, Hanzi and Rolland, Chloe and Gustafson, Laura and Xiao, Tete and Whitehead, Spencer and Berg, Alexander C. and Lo, Wan-Yen and Doll{\'a}r, Piotr and Girshick, Ross},
  journal={arXiv:2304.02643},
  year={2023}
}

@article{ravi2024sam2,
  title={SAM 2: Segment Anything in Images and Videos},
  author={Ravi, Nikhila and Gabeur, Valentin and Hu, Yuan-Ting and Hu, Ronghang and Ryali, Chaitanya and Ma, Tengyu and Khedr, Haitham and R{\"a}dle, Roman and Rolland, Chloe and Gustafson, Laura and Mintun, Eric and Pan, Junting and Alwala, Kalyan Vasudev and Carion, Nicolas and Wu, Chao-Yuan and Girshick, Ross and Doll{\'a}r, Piotr and Feichtenhofer, Christoph},
  journal={arXiv preprint arXiv:2408.00714},
  url={https://arxiv.org/abs/2408.00714},
  year={2024}
}

@misc{carion2025sam3segmentconcepts,
      title={SAM 3: Segment Anything with Concepts},
      author={Nicolas Carion and Laura Gustafson and Yuan-Ting Hu and Shoubhik Debnath and Ronghang Hu and Didac Suris and Chaitanya Ryali and Kalyan Vasudev Alwala and Haitham Khedr and Andrew Huang and Jie Lei and Tengyu Ma and Baishan Guo and Arpit Kalla and Markus Marks and Joseph Greer and Meng Wang and Peize Sun and Roman Rädle and Triantafyllos Afouras and Effrosyni Mavroudi and Katherine Xu and Tsung-Han Wu and Yu Zhou and Liliane Momeni and Rishi Hazra and Shuangrui Ding and Sagar Vaze and Francois Porcher and Feng Li and Siyuan Li and Aishwarya Kamath and Ho Kei Cheng and Piotr Dollár and Nikhila Ravi and Kate Saenko and Pengchuan Zhang and Christoph Feichtenhofer},
      year={2025},
      eprint={2511.16719},
      archivePrefix={arXiv},
      primaryClass={cs.CV},
      url={https://arxiv.org/abs/2511.16719},
}

@inproceedings{wu2024ppt,
    title={Towards Large-scale 3D Representation Learning with Multi-dataset Point Prompt Training},
    author={Wu, Xiaoyang and Tian, Zhuotao and Wen, Xin and Peng, Bohao and Liu, Xihui and Yu, Kaicheng and Zhao, Hengshuang},
    booktitle={CVPR},
    year={2024}
}

@misc{pointcept2023,
    title={Pointcept: A Codebase for Point Cloud Perception Research},
    author={Pointcept Contributors},
    howpublished={\url{https://github.com/Pointcept/Pointcept}},
    year={2023}
}

@INPROCEEDINGS {randlanet2020,
author = { Hu, Qingyong and Yang, Bo and Xie, Linhai and Rosa, Stefano and Guo, Yulan and Wang, Zhihua and Trigoni, Niki and Markham, Andrew },
booktitle = { 2020 IEEE/CVF Conference on Computer Vision and Pattern Recognition (CVPR) },
title = {{ RandLA-Net: Efficient Semantic Segmentation of Large-Scale Point Clouds }},
year = {2020},
volume = {},
ISSN = {},
pages = {11105-11114},
doi = {10.1109/CVPR42600.2020.01112},
url = {https://doi.ieeecomputersociety.org/10.1109/CVPR42600.2020.01112},
publisher = {IEEE Computer Society},
address = {Los Alamitos, CA, USA},
month =Jun}

@misc{saccon2026combininglargelanguagemodels,
      title={Combining Large Language Models and Symbolic Reasoning for Multi-Robot Temporal Planning through Explainable Knowledge Bases}, 
      author={Enrico Saccon and Matteo Saveriano and Edoardo Lamon and Luigi Palopoli and Marco Roveri},
      year={2026},
      eprint={2502.19135},
      archivePrefix={arXiv},
      primaryClass={cs.AI},
      url={https://arxiv.org/abs/2502.19135}, 
}

@article{saccon2025automated,
  title={Automated generation of mdps using logic programming and llms for robotic applications},
  author={Saccon, Enrico and De Martini, Davide and Saveriano, Matteo and Lamon, Edoardo and Palopoli, Luigi and Roveri, Marco},
  journal={IEEE Robotics and Automation Letters},
  volume={11},
  number={2},
  pages={1770--1777},
  year={2025},
  publisher={IEEE}
}

@misc{singh2026openaigpt5card,
      title={OpenAI GPT-5 System Card}, 
      author={Aaditya Singh and Adam Fry and Adam Perelman and Adam Tart and Adi Ganesh and Ahmed El-Kishky and Aidan McLaughlin and Aiden Low and AJ Ostrow and Akhila Ananthram and Akshay Nathan and Alan Luo and Alec Helyar and Aleksander Madry and Aleksandr Efremov and Aleksandra Spyra and Alex Baker-Whitcomb and Alex Beutel and Alex Karpenko and Alex Makelov and Alex Neitz and Alex Wei and Alexandra Barr and Alexandre Kirchmeyer and Alexey Ivanov and Alexi Christakis and Alistair Gillespie and Allison Tam and Ally Bennett and Alvin Wan and Alyssa Huang and Amy McDonald Sandjideh and Amy Yang and Ananya Kumar and Andre Saraiva and Andrea Vallone and Andrei Gheorghe and Andres Garcia Garcia and Andrew Braunstein and Andrew Liu and Andrew Schmidt and Andrey Mereskin and Andrey Mishchenko and Andy Applebaum and Andy Rogerson and Ann Rajan and Annie Wei and Anoop Kotha and Anubha Srivastava and Anushree Agrawal and Arun Vijayvergiya and Ashley Tyra and Ashvin Nair and Avi Nayak and Ben Eggers and Bessie Ji and Beth Hoover and Bill Chen and Blair Chen and Boaz Barak and Borys Minaiev and Botao Hao and Bowen Baker and Brad Lightcap and Brandon McKinzie and Brandon Wang and Brendan Quinn and Brian Fioca and Brian Hsu and Brian Yang and Brian Yu and Brian Zhang and Brittany Brenner and Callie Riggins Zetino and Cameron Raymond and Camillo Lugaresi and Carolina Paz and Cary Hudson and Cedric Whitney and Chak Li and Charles Chen and Charlotte Cole and Chelsea Voss and Chen Ding and Chen Shen and Chengdu Huang and Chris Colby and Chris Hallacy and Chris Koch and Chris Lu and Christina Kaplan and Christina Kim and CJ Minott-Henriques and Cliff Frey and Cody Yu and Coley Czarnecki and Colin Reid and Colin Wei and Cory Decareaux and Cristina Scheau and Cyril Zhang and Cyrus Forbes and Da Tang and Dakota Goldberg and Dan Roberts and Dana Palmie and Daniel Kappler and Daniel Levine and Daniel Wright and Dave Leo and David Lin and David Robinson and Declan Grabb and Derek Chen and Derek Lim and Derek Salama and Dibya Bhattacharjee and Dimitris Tsipras and Dinghua Li and Dingli Yu and DJ Strouse and Drew Williams and Dylan Hunn and Ed Bayes and Edwin Arbus and Ekin Akyurek and Elaine Ya Le and Elana Widmann and Eli Yani and Elizabeth Proehl and Enis Sert and Enoch Cheung and Eri Schwartz and Eric Han and Eric Jiang and Eric Mitchell and Eric Sigler and Eric Wallace and Erik Ritter and Erin Kavanaugh and Evan Mays and Evgenii Nikishin and Fangyuan Li and Felipe Petroski Such and Filipe de Avila Belbute Peres and Filippo Raso and Florent Bekerman and Foivos Tsimpourlas and Fotis Chantzis and Francis Song and Francis Zhang and Gaby Raila and Garrett McGrath and Gary Briggs and Gary Yang and Giambattista Parascandolo and Gildas Chabot and Grace Kim and Grace Zhao and Gregory Valiant and Guillaume Leclerc and Hadi Salman and Hanson Wang and Hao Sheng and Haoming Jiang and Haoyu Wang and Haozhun Jin and Harshit Sikchi and Heather Schmidt and Henry Aspegren and Honglin Chen and Huida Qiu and Hunter Lightman and Ian Covert and Ian Kivlichan and Ian Silber and Ian Sohl and Ibrahim Hammoud and Ignasi Clavera and Ikai Lan and Ilge Akkaya and Ilya Kostrikov and Irina Kofman and Isak Etinger and Ishaan Singal and Jackie Hehir and Jacob Huh and Jacqueline Pan and Jake Wilczynski and Jakub Pachocki and James Lee and James Quinn and Jamie Kiros and Janvi Kalra and Jasmyn Samaroo and Jason Wang and Jason Wolfe and Jay Chen and Jay Wang and Jean Harb and Jeffrey Han and Jeffrey Wang and Jennifer Zhao and Jeremy Chen and Jerene Yang and Jerry Tworek and Jesse Chand and Jessica Landon and Jessica Liang and Ji Lin and Jiancheng Liu and Jianfeng Wang and Jie Tang and Jihan Yin and Joanne Jang and Joel Morris and Joey Flynn and Johannes Ferstad and Johannes Heidecke and John Fishbein and John Hallman and Jonah Grant and Jonathan Chien and Jonathan Gordon and Jongsoo Park and Jordan Liss and Jos Kraaijeveld and Joseph Guay and Joseph Mo and Josh Lawson and Josh McGrath and Joshua Vendrow and Joy Jiao and Julian Lee and Julie Steele and Julie Wang and Junhua Mao and Kai Chen and Kai Hayashi and Kai Xiao and Kamyar Salahi and Kan Wu and Karan Sekhri and Karan Sharma and Karan Singhal and Karen Li and Kenny Nguyen and Keren Gu-Lemberg and Kevin King and Kevin Liu and Kevin Stone and Kevin Yu and Kristen Ying and Kristian Georgiev and Kristie Lim and Kushal Tirumala and Kyle Miller and Lama Ahmad and Larry Lv and Laura Clare and Laurance Fauconnet and Lauren Itow and Lauren Yang and Laurentia Romaniuk and Leah Anise and Lee Byron and Leher Pathak and Leon Maksin and Leyan Lo and Leyton Ho and Li Jing and Liang Wu and Liang Xiong and Lien Mamitsuka and Lin Yang and Lindsay McCallum and Lindsey Held and Liz Bourgeois and Logan Engstrom and Lorenz Kuhn and Louis Feuvrier and Lu Zhang and Lucas Switzer and Lukas Kondraciuk and Lukasz Kaiser and Manas Joglekar and Mandeep Singh and Mandip Shah and Manuka Stratta and Marcus Williams and Mark Chen and Mark Sun and Marselus Cayton and Martin Li and Marvin Zhang and Marwan Aljubeh and Matt Nichols and Matthew Haines and Max Schwarzer and Mayank Gupta and Meghan Shah and Melody Y. Guan and Melody Huang and Meng Dong and Mengqing Wang and Mia Glaese and Micah Carroll and Michael Lampe and Michael Malek and Michael Sharman and Michael Zhang and Michele Wang and Michelle Pokrass and Mihai Florian and Mikhail Pavlov and Miles Wang and Ming Chen and Mingxuan Wang and Minnia Feng and Mo Bavarian and Molly Lin and Moose Abdool and Mostafa Rohaninejad and Nacho Soto and Natalie Staudacher and Natan LaFontaine and Nathan Marwell and Nelson Liu and Nick Preston and Nick Turley and Nicklas Ansman and Nicole Blades and Nikil Pancha and Nikita Mikhaylin and Niko Felix and Nikunj Handa and Nishant Rai and Nitish Keskar and Noam Brown and Ofir Nachum and Oleg Boiko and Oleg Murk and Olivia Watkins and Oona Gleeson and Pamela Mishkin and Patryk Lesiewicz and Paul Baltescu and Pavel Belov and Peter Zhokhov and Philip Pronin and Phillip Guo and Phoebe Thacker and Qi Liu and Qiming Yuan and Qinghua Liu and Rachel Dias and Rachel Puckett and Rahul Arora and Ravi Teja Mullapudi and Raz Gaon and Reah Miyara and Rennie Song and Rishabh Aggarwal and RJ Marsan and Robel Yemiru and Robert Xiong and Rohan Kshirsagar and Rohan Nuttall and Roman Tsiupa and Ronen Eldan and Rose Wang and Roshan James and Roy Ziv and Rui Shu and Ruslan Nigmatullin and Saachi Jain and Saam Talaie and Sam Altman and Sam Arnesen and Sam Toizer and Sam Toyer and Samuel Miserendino and Sandhini Agarwal and Sarah Yoo and Savannah Heon and Scott Ethersmith and Sean Grove and Sean Taylor and Sebastien Bubeck and Sever Banesiu and Shaokyi Amdo and Shengjia Zhao and Sherwin Wu and Shibani Santurkar and Shiyu Zhao and Shraman Ray Chaudhuri and Shreyas Krishnaswamy and Shuaiqi and Xia and Shuyang Cheng and Shyamal Anadkat and Simón Posada Fishman and Simon Tobin and Siyuan Fu and Somay Jain and Song Mei and Sonya Egoian and Spencer Kim and Spug Golden and SQ Mah and Steph Lin and Stephen Imm and Steve Sharpe and Steve Yadlowsky and Sulman Choudhry and Sungwon Eum and Suvansh Sanjeev and Tabarak Khan and Tal Stramer and Tao Wang and Tao Xin and Tarun Gogineni and Taya Christianson and Ted Sanders and Tejal Patwardhan and Thomas Degry and Thomas Shadwell and Tianfu Fu and Tianshi Gao and Timur Garipov and Tina Sriskandarajah and Toki Sherbakov and Tomek Korbak and Tomer Kaftan and Tomo Hiratsuka and Tongzhou Wang and Tony Song and Tony Zhao and Troy Peterson and Val Kharitonov and Victoria Chernova and Vineet Kosaraju and Vishal Kuo and Vitchyr Pong and Vivek Verma and Vlad Petrov and Wanning Jiang and Weixing Zhang and Wenda Zhou and Wenlei Xie and Wenting Zhan and Wes McCabe and Will DePue and Will Ellsworth and Wulfie Bain and Wyatt Thompson and Xiangning Chen and Xiangyu Qi and Xin Xiang and Xinwei Shi and Yann Dubois and Yaodong Yu and Yara Khakbaz and Yifan Wu and Yilei Qian and Yin Tat Lee and Yinbo Chen and Yizhen Zhang and Yizhong Xiong and Yonglong Tian and Young Cha and Yu Bai and Yu Yang and Yuan Yuan and Yuanzhi Li and Yufeng Zhang and Yuguang Yang and Yujia Jin and Yun Jiang and Yunyun Wang and Yushi Wang and Yutian Liu and Zach Stubenvoll and Zehao Dou and Zheng Wu and Zhigang Wang},
      year={2026},
      eprint={2601.03267},
      archivePrefix={arXiv},
      primaryClass={cs.CL},
      url={https://arxiv.org/abs/2601.03267}, 
}

@misc{gestrin2025nl2planrobustllmdrivenplanning,
      title={NL2Plan: Robust LLM-Driven Planning from Minimal Text Descriptions}, 
      author={Elliot Gestrin and Marco Kuhlmann and Jendrik Seipp},
      year={2025},
      eprint={2405.04215},
      archivePrefix={arXiv},
      primaryClass={cs.AI},
      url={https://arxiv.org/abs/2405.04215}, 
}

@misc{zhao2023fastsegment,
  title={Fast Segment Anything}, 
  author={Xu Zhao and Wenchao Ding and Yongqi An and Yinglong Du and Tao Yu and Min Li and Ming Tang and Jinqiao Wang},
  year={2023},
  eprint={2306.12156},
  archivePrefix={arXiv},
  primaryClass={cs.CV},
  url={https://arxiv.org/abs/2306.12156}, 
}

@misc{dataset,
    title = {{RGB-D Dataset for Object-Centric Robotic Scene Understanding}},
    author = {Enrico Saccon and Tommaso Faraci and {I\~{n}igo} {De La Ossa Zarzuelo} and Luigi Palopoli and Marco Roveri and Matteo Saveriano},
    year = {2026},
    publisher = {Zenodo},
    url = {doi.org/10.5281/zenodo.22902240}
}

\end{document}